\documentclass{article}

\usepackage[numbers, compress]{natbib}

\usepackage[utf8]{inputenc} 
\usepackage[T1]{fontenc}    
\usepackage{hyperref}       
\usepackage{url}            
\usepackage{booktabs}       
\usepackage{amsfonts}       
\usepackage{nicefrac}       
\usepackage{microtype}      
\usepackage{xcolor}         
\usepackage{graphicx}
\usepackage{amsmath}
\usepackage[ruled,vlined]{algorithm2e}
\usepackage{verbatim}
\usepackage{geometry}

\title{Evolving Computation Graphs}

\author{%
  Andreea Deac \\
  Mila – Qu{\'e}bec AI Institute \\ Universit{\'e} de Montr{\'e}al \\
  \and
  Jian Tang \\
  Mila – Qu{\'e}bec AI Institute \\ HEC Montr{\'e}al
}

\date{}

\begin{document}

\maketitle

\begin{abstract}
  Graph neural networks (GNNs) have demonstrated success in modeling relational data, especially for data that exhibits homophily: when a connection between nodes tends to imply that they belong to the same class. However, while this assumption is true in many relevant situations, there are important real-world scenarios that violate this assumption, and this has spurred research into improving GNNs for these cases. In this work, we propose Evolving Computation Graphs (ECGs), a novel method for enhancing GNNs on heterophilic datasets. Our approach builds on prior theoretical insights linking node degree, high homophily, and inter vs intra-class embedding similarity by rewiring the GNNs’ computation graph towards adding edges that connect nodes that are likely to be in the same class. We utilise weaker classifiers to identify these edges, ultimately improving GNN performance on non-homophilic data as a result. We evaluate ECGs on a diverse set of recently-proposed heterophilous datasets and demonstrate improvements over the relevant baselines. ECG presents a simple, intuitive and elegant approach for improving GNN performance on heterophilic datasets without requiring prior domain knowledge.
\end{abstract}

\section{Introduction}
Neural networks applied to graph-structured data have demonstrated success across various domains, including practical applications like drug discovery \cite{stokes2020deep}, transportation networks \cite{derrow2021eta}, chip design \cite{mirhoseini2021graph} and theoretical advancements \cite{davies2021advancing, blundell2021towards}. Numerous architectures fall under the category of graph neural networks \cite{bronstein2021geometric}, with one of the most versatile ones being Message Passing Neural Networks \cite{gilmer2017neural}. The fundamental concept behind these networks is that nodes communicate with their neighbouring nodes through messages in each layer. These messages, received from neighbours, are then aggregated in a permutation-invariant manner to contribute to a new node representation.

It has been observed that the performance of graph neural networks may rely on the underlying assumption of \textit{homophily}, which suggests that nodes are connected by edges if they are similar based on their attributes or belonging to the same class, as commonly seen in social or citation networks. However, this assumption often fails to accurately describe real-world data when the graph contains \textit{heterophilic} edges, connecting dissimilar nodes. This observation holds particular significance since graph neural networks tend to exhibit significantly poorer performance on heterophilic graphs compared to datasets known to be homophilic. Several studies \cite{zhu2020beyond, zhu2020graph, wang2022powerful, he2022block} have highlighted this issue, using a mixture of strongly homophilous graphs---such as Cora, Citeseer and Pubmed \cite{sen2008collective}---as well a standard suite of six heterophilic datasets---Squirrel, Chameleon, Cornell, Texas, Wisconsin and Actor \cite{rozemberczki2021multi,tang2009social}---first curated jointly by \citet{pei2020geom}. 

\begin{figure}
    \centering
    \includegraphics[width=0.6\linewidth]{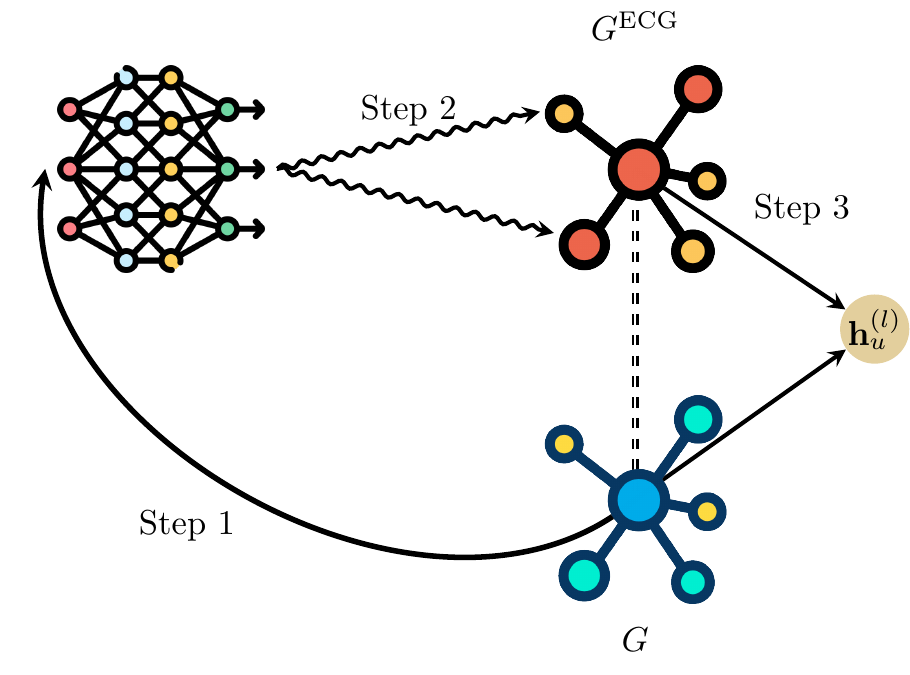}
    \caption{A simplified illustration of Evolving Computation Graphs. {\bf Step 1:} nodes in a graph, $G$, are embedded using a pre-trained weak classifier. {\bf Step 2:} Based on these embeddings, a nearest-neighbour graph, $G^\mathrm{EGC}$, is generated. This graph is likely to have improved propagation and homophily properties (illustrated by similar colours between neighbouring nodes). {\bf Step 3:} Message passing is performed, both in the original and in the ECG graph, to update node representations.}
    \label{fig:first_fig}
\end{figure}

In the context of this standard suite of heterophilic graphs, it has been observed that general graph neural network architectures tend to underperform unless there is high label informativeness \cite{ma2021homophily, platonov2022characterizing}. In prior work, this issue was tackled primarily by proposing modifications to the GNN architecture. These modifications include changes to the aggregation function, such as separating self- and neighbour embeddings \cite{zhu2020beyond}, mixing low- and high-frequency signals \cite{bo2021beyond, luan2022revisiting}, and predicting and utilising the compatibility matrix \cite{zhu2020generalizing}. Other approaches involve using the Jacobi basis in spectral GNNs \cite{wang2022powerful} or learning cellular sheaves for neural sheaf diffusion \cite{bodnar2022neural} to improve performance.

However, it was recently remarked \cite{platonov2023critical} that this standard heterophilous suite has significant drawbacks, such as data originating from only three sources, two of the datasets having significant numbers of repeated nodes and improper evaluation regarding class imbalance. 
To address these shortcomings, a more recent benchmark suite has been introduced by \citet{platonov2023critical}, incorporating improvements on all of the above issues.
Interestingly, once such corrections are accounted for, standard GNN models such as graph convolutional networks \cite[GCN]{kipf2016classification}, GraphSAGE \cite[SAGE]{hamilton2017inductive}, graph attention networks \cite[GAT]{velivckovic2017attention}, and Graph Transformers \cite[GT]{shi2020masked} have demonstrated superior performance compared to specialized architectures tailored specifically for heterophily---in spite of the heterophilic properties of the datasets. The notable exception is \textit{-sep} \cite{zhu2020beyond} which has consistently improved GAT and GT by modelling self and neighbouring nodes separately. 

In light of this surprising discovery, we suggest that there should be alternate routes to making the most of heterophilic datasets. Rather than attempting to modify these standard GNNs, we propose modifying their \emph{computation graph}: effectively, enforcing messages to be sent across additional pairs of nodes. These node pairs are chosen according to a particular measure of \emph{similarity}. If the similarity metric is favourably chosen, such a computation graph will improve the overall homophily statistics, thereby creating more favourable conditions for GNNs to perform well.

We further propose that the \emph{modification} of the computation graph should be separate from its \emph{utilisation}\footnote{We note that \cite{pei2020geom,suresh2021breaking} also rest on a similar proposal, however their updated computation graph is fully derived as a function of the input graph structure (ignoring node features and labels), and thus it is unavoidably \emph{vulnerable} to any biases or inconsistencies in the input graph---and real-world graph inputs are rarely flawless.}. That is, we proceed in two phases: the first phase learns the representations that allow us to construct new computation graphs, and the second phase utilises those representations to construct new computation graphs, to be utilised by a GNN in each layer. This design choice makes our method elegant, performant and easy to evaluate: the two-phase nature means we are not susceptible to bilevel optimisation (as in \cite{franceschi2019learning,kazi2022differentiable}), the graphs we use need to be precomputed exactly once rather than updated on-the-fly in every layer (as in \cite{wang2019dynamic}), and because the same computation graph is used across all GNN layers, we can more rigidly evaluate how useful this graph is, all other things kept equal.

Hence, the essence of our method is \emph{Evolving Computation Graphs} (\textbf{ECG}), which uses weak classifiers to generate node embeddings. These embeddings are then used to define a similarity metric between nodes (such as cosine similarity). We then select edges in a $k$-nearest neighbour fashion: we connect each node to $k$ nodes most similar to it, according to the metric. The edges selected in this manner form a complementary graph, which we propose using in parallel with the input graph to update each node's representation. For this purpose, we use standard, off-the-shelf, GNNs. Our method is illustrated in Figure \ref{fig:first_fig}.

The nature of the weak classifier employed in ECG is flexible and, for the purpose of this paper, we used two representative options. The first option is a point-wise MLP classifier, attempting to cluster together nodes based on the given training labels, without any graph-based biases. For the second option, we attempt the converse: utilising the given graph structure and node features, but not relying on the training labels. This is a suitable setting for a self-supervised graph representation learning method, such as BGRL \cite{thakoor2021large}, which is designed to cluster together nodes with similar local neighbourhoods---both in terms of subgraphs and features---through a bootstrapping objective \cite{grill2020bootstrap}. 

To evaluate the effectiveness of ECG, we conduct experiments on the benchmark suite proposed by \citet{platonov2023critical}. Our results demonstrate that ECG models outperform their GNN baselines in 19 out of 20 head-to-head comparisons. The most significant improvements can be noticed for GCNs---which are best suited to benefit from improved homophily---where improvements reach up to $10\%$ in absolute terms. Further, the best performing ECG models outperform a diverse set of representative heterophily-specialised GNNs.

\section{Background}

In this section, we introduce the generic setup of learning representations on graphs, along with all of the key components that comprise our ECG method.

\paragraph{Graph representation learning setup} We denote graphs by $G=(V,E)$, where $V$ is the set of nodes and $E$ is the set of edges, and we denote by $e_{uv}\in E$ the edge that connects nodes $u$ and $v$. For the datasets considered here, we can assume that the input graphs are provided to the GNNs via two inputs: the \emph{node feature matrix}, ${\bf X}\in\mathbb{R}^{|V|\times k}$ (such that $\mathbf{x}_u\in\mathbb{R}^k$ are the input features of node $u\in V$), and the \emph{adjacency matrix}, ${\bf A}\in\{0, 1\}^{|V|\times |V|}$, such that $a_{uv}$ indicates whether nodes $u$ and $v$ are connected by an edge. We further assume the graph is \emph{undirected}; that is, ${\bf A} = {\bf A}^\top$. We also use $d_u = \sum_{v\in V} a_{uv} \left(= \sum_{v\in V} a_{vu}\right)$ to denote the degree of node $u$.

We focus on node classification tasks with $C$ representing the set of possible classes, where for node with input features $\mathbf{x}_u$, there is a label $y_u \in C$. Thus we aim to learn a function $f$ that minimises $\mathbb{E}[\mathcal{L}(y_u, \hat{y}_u)]$, where $ \hat{y}_u$ is the prediction of $f(\mathbf{x}_u) = \hat{y}_u$, and $\mathcal{L}$ is the cross-entropy loss function.

\paragraph{Graph neural networks} The one-step layer of a GNN can be summarised as follows \cite{bronstein2021geometric}:
\begin{equation}\label{eq:mpnn}
    \mathbf{h}_u^{(l)} = \phi^{(l)}\left( \mathbf{h}_u^{(l-1)}, \bigoplus_{(u,v) \in E} \psi^{(l)}\left(\mathbf{h}_u^{(l-1)}, \mathbf{h}_v^{(l-1)}\right)\right)
\end{equation}
where, by definition, we set $\mathbf{h}_u^{(0)} = \mathbf{x}_u$. Leveraging different (potentially learnable) functions for $\phi^{(l)} : \mathbb{R}^k\times\mathbb{R}^m\rightarrow\mathbb{R}^{k'}$, $\bigoplus : \mathrm{bag}(\mathbb{R}^m)\rightarrow\mathbb{R}^m$ and $\psi^{(l)} : \mathbb{R}^k\times\mathbb{R}^k\rightarrow\mathbb{R}^m$ then recovers well-known GNN architectures. Examples include GCN \cite{kipf2016semi}: $\psi^{(l)}(\mathbf{x}_u, \mathbf{x}_v)=\beta_{uv} \omega^{(l)}(\mathbf{x}_v)$, with $\beta_{uv}\in\mathbb{R}$ being a constant based on ${\bf A}$, GAT \cite{velivckovic2017graph}: $\psi^{(l)}(\mathbf{x}_u, \mathbf{x}_v)=\alpha^{(l)}(\mathbf{x}_u, \mathbf{x}_v) \omega^{(l)}(\mathbf{x}_v)$ with $\alpha^{(l)} : \mathbb{R}^k\times\mathbb{R}^k\rightarrow\mathbb{R}$ being a (softmax-normalised) self-attention mechanism, and \textit{-sep} \cite{zhu2020beyond}: $\phi^{(l)} = {\bf W}^{(l)}_\mathrm{self}\phi^{(l)}_1\left(\mathbf{h}_u^{(l-1)}\right) + {\bf W}^{(l)}_\mathrm{agg}\phi^{(l)}_2\left(\bigoplus_{(u,v) \in E} \psi^{(l)}(\mathbf{h}_u^{(l-1)}, \mathbf{h}_v^{(l-1)})\right)$, where we explicitly decompose $\phi^{(l)}$ into two parts, with one of them ($\phi^{(l)}_1$) depending on the receiver node only.

\paragraph{Homophily} has been repeatedly mentioned as an important measure of the graph, especially when it comes to GNN performance. Intuitively, it corresponds to an assumption that neighbouring nodes tend to share labels: $a_{uv} = 1 \implies y_u = y_v$, which is often the case for many industrially-relevant real world graphs (such as social networks). Intuitively, a graph with high homophily will make it easier to exploit neighbourhood structure to derive more accurate node labels.

However, in spite of the importance of quantifying homophily in a graph, there is no universally-agreed-upon metric for this. One very popular metric, used by several studies, is \textit{edge homophily} \cite{abu2019mixhop}, which measures the proportion of homophilic edges:
\begin{equation}
\texttt{h-edge} = \frac{|(u,v) \in E: y_u = y_v|}{|E|}
\end{equation}
while \cite{platonov2022characterizing} also introduces \textit{adjusted homophily} to account for number of classes and their distributions:
\begin{equation}
\texttt{h-adj} = \frac{\texttt{h-edge} - \sum_{k=1}^C D_k^2 / (2|E|)^2 }{ 1- \sum_{k=1}^C D_k^2 / (2|E|)^2}
\end{equation}
where $D_k = \sum_{u:y_u=k} d_u$, the sum of degrees for the nodes belonging to class $k$.

Additionally, the \textit{label informativeness} (LI) measure proposed in \cite{platonov2022characterizing} measures how much information about a node's label is gained by observing its neighbour's label, on average. It is defined as 
\begin{equation}
    \texttt{LI} = I(y_{\xi}, y_{\eta}) / H(y_{\xi})
\end{equation}
where $(\xi, \eta) \in E$ is a uniformly-sampled edge, $H$ is the Shannon entropy and $I$ is mutual information.

\paragraph{Weak classifier} In order to derive novel computation graphs which are likely to result in higher test performance, we likely require ``novel'' homophilic connections to emerge---rather than amplifying the homophily already present in ${\bf A}$. Therefore, for the purposes of building a useful computation graph, our ECG method aims to first learn representations of nodes governed by a model which does \emph{not} have access to inputs (${\bf X}$), graph structure (${\bf A}$) and training labels (${\bf y}_\mathrm{tr}$) simultaneously. We hence call such a model a ``weak classifier'', as it is not exposed to the same kind of inductive biases as a supervised GNN would (and hence it must obtain useful models which do not rely on these biases).

\paragraph{MLPs} Arguably the simplest way to make a weak classifier, as above, is to withhold access to the graph structure (${\bf A}$), and force the model to classify the nodes in pure isolation from one another. This is effectively a standard multi-layer perceptron (MLP) applied pointwise. Another way of understanding this model is setting ${\bf A} = {\bf I}_{|V|}$, or equivalently, $E = \{(u, u)\ |\ u\in V\}$, in Equation \ref{eq:mpnn}, which is sometimes referred to as the Deep Sets model \citep{zaheer2017deep}. We train this model by using cross-entropy against the training nodes' labels (${\bf y}_\mathrm{tr}$) and, once trained, use the final layer activations, $\mathbf{h}_u^{(L)}$---for a model with $L$ layers---as our MLP embeddings.

\paragraph{BGRL} While using the embeddings from an MLP can offer a solid way to improve homophily metrics, their confidence will degrade for nodes where the model is less accurate outside of the training set---which are arguably the nodes we would like to improve predictions on the most. Accordingly, as a converse approach to obtaining a weak classifier, we may also withhold access to the training labels ($\mathbf{y}_\mathrm{tr}$). Now the model is forced to arrange the node representations in a way that will be mindful of the input features and graph structure, but without knowing the task specifics upfront, and hence not vulnerable to overfitting on the training nodes. Such a weak classifier naturally lends itself to self-supervised learning on graphs.

Bootstrapped graph latents \cite[BGRL]{thakoor2021large} is a state-of-the-art self-supervised graph representation learning method based on BYOL \cite{grill2020bootstrap}. BGRL learns two GNN encoders with identical architecture; an \emph{online} encoder, $\mathcal{E}_\theta$, and a \emph{target} encoder, $\mathcal{E}_\phi$. BGRL also contains a \emph{predictor} network $p_\theta$. We offer a ``bird's eye'' view of how BGRL is trained, and defer to \cite{thakoor2021large} for implementation details.

At each step of training, BGRL proceeds as follows. First, two data augmentations (e.g. random node and edge dropout) are applied to the input graph, obtaining augmented graphs $({\bf X}_1, {\bf A}_1)$ and $({\bf X}_2, {\bf A}_2)$. Then, the two encoders are applied to these augmentations, recovering a pair of latent node embeddings: ${\bf H}_1 = \mathcal{E}_\theta({\bf X}_1, {\bf A}_1)$, ${\bf H}_2 = \mathcal{E}_\phi({\bf X}_2, {\bf A}_2)$. The first embedding is additionally passed through the predictor network: ${\bf Z}_1 = p_\theta({\bf H}_1)$. At this point, BGRL attempts to preserve the cosine similarity between all the corresponding nodes in ${\bf Z}_1$ and ${\bf H}_2$, via the following loss function:
\begin{equation}
\label{eqn:bgrl_loss}
    \mathcal{L}_{\mathrm{BGRL}} = -\frac{{\bf Z}_1{\bf H}_2^\top}{\|{\bf Z}_1\| \|{\bf H}_2\|}
\end{equation}
Lastly, the parameters of the online encoder $\mathcal{E}_\theta$ and predictor $p_\theta$ are updated via stochastic gradient descent on $\mathcal{L}_\mathrm{BGRL}$, and the parameters of the target encoder $\mathcal{E}_\phi$ are updated as the exponential moving average of the online encoder's parameters.

Once the training procedure concludes, typically only the online network $\mathcal{E}_\theta$ is retained, and hence the embeddings ${\bf H} = \mathcal{E}_\theta({\bf X}, {\bf A})$ are the BGRL embeddings of the input graph given by node features ${\bf X}$ and adjacency matrix ${\bf A}$.

Owing to its bootstrapped objective, BGRL does not require the generation of negative samples, and is hence computationally efficient compared to contrastive learning approaches. Further, it is very successful at large scales; it was shown by \citet{addanki2021large} that BGRL's benefits persist on industrially relevant graphs of hundreds of millions of nodes, leading to one of the top-3 winning entries at the OGB-LSC competition \citep{hu2021ogb}. This is why we employ it as a representative self-supervised embedding method for our ECG framework.

\section{Evolving Computation Graphs}\label{sec:ECG}
Armed with the concepts above, we are now ready to describe the steps of the ECG methodology. Please refer to Algorithm \ref{algo:method} for a pseudocode summary.

\paragraph{Step 1: Embedding extraction}

Firstly, we assume that an appropriate weak classifier has already been trained (as discussed in previous sections), and is capable of producing node embeddings. We start by invoking this classifier to obtain ECG embeddings $\mathbf{H}_\mathrm{ECG} = \gamma(\mathbf{X}, \mathbf{A})$. We study two simple but potent variants of $\gamma$, as per the previous section:

\begin{itemize}
    \item[\bf MLP:] In this case, we utilise a simple deep MLP\footnote{Note that this MLP only computes high-dimensional embeddings of each node; while training $\gamma$, an additional logistic regression layer is attached to this architecture.}; that is, $\gamma(\mathbf{X}, \mathbf{A}) = \sigma\left(\sigma\left(\mathbf{X}\mathbf{W}_1\right)\mathbf{W}_2\right)$, where $\mathbf{W}_\cdot$ are the weights of the MLP, and $\sigma$ is the GELU activation function \citep{hendrycks2016gaussian}.
    \item[\bf BGRL:] In this case, we set $\gamma = \mathcal{E}_\theta$, the online encoder of BGRL. For our experiments, we utilise a publicly available off-the-shelf implementation of BGRL provided by the Deep Graph Library\footnote{\url{https://github.com/dmlc/dgl/tree/master/examples/pytorch/bgrl}} \citep{wang2019dgl}, which uses a two-layer GCN \cite{kipf2016semi} as the base encoder.
\end{itemize}
The parameters of $\gamma$ are kept frozen throughout, and are not to be further trained on.

\paragraph{Step 2: Graph construction}
Having obtained $\mathbf{H}_\mathrm{ECG}$, we can now use it to compute a similarity metric between the nodes, such as cosine similarity, as follows: 
\begin{equation}
    \mathbf{S} = \mathbf{H}_\mathrm{ECG}\mathbf{H}_\mathrm{ECG}^\top \qquad \hat{s}_{uv} = \frac{s_{uv}}{\|\mathbf{h}_{\mathrm{ECG}_u}\|\|\mathbf{h}_{\mathrm{ECG}_v}\|} 
\end{equation}

Based on this similarity metric, for each node $u\in V$ we select its neighbourhood $\mathcal{N}_u^\mathrm{ECG}$ to be its $k$ nearest neighbours in $\mathbf{S}$ (where $k$ is a tunable hyperparameter):
\begin{equation}
    \mathcal{N}_u^\mathrm{ECG} = \mathrm{top\text{-}}k_{v\in V} \hat{s}_{uv}
\end{equation}

Equivalently, we construct a new computation graph, $G^{\mathrm{ECG}}=(V, E^\mathrm{ECG})$, such that its edges are $E^\mathrm{ECG} = \{(u, v)\ |\ u\in V\wedge v\in\mathcal{N}_u\}$. These edges are effectively determined by the weak classifier.

\paragraph{Step 3: Parallel message passing} Finally, once the ECG graph, $G^\mathrm{ECG}$, is available, we can run our GNN of choice over it. To retain the topological benefits contained in the input graph structure, we opt to run two GNN layers in parallel---one over the input graph (as in Equation \ref{eq:mpnn}), and one over the ECG graph, as follows:
\begin{align}\label{eqn:tu}
    \mathbf{h}_{\mathrm{INP}_u}^{(l)} &= \phi^{(l)}_\mathrm{INP}\left( \mathbf{h}_u^{(l-1)}, \bigoplus_{(u,v) \in E} \psi_\mathrm{INP}^{(l)}\left(\mathbf{h}_u^{(l-1)}, \mathbf{h}_v^{(l-1)}\right)\right)\\
    \mathbf{h}_{\mathrm{ECG}_u}^{(l)} &= \phi^{(l)}_\mathrm{ECG}\left( \mathbf{h}_u^{(l-1)}, \bigoplus_{(u,v) \in E^\mathrm{ECG}} \psi_\mathrm{ECG}^{(l)}\left(\mathbf{h}_u^{(l-1)}, \mathbf{h}_v^{(l-1)}\right)\right)
\end{align}

Then the representation after $l$ layers is obtained by jointly transforming these two representations:
\begin{equation}\label{eqn:te}
    \mathbf{h}_{u}^{(l)} = \mathbf{W}^{(l)} \mathbf{h}_{\mathrm{INP}_u}^{(l)} + \mathbf{U}^{(l)} \mathbf{h}_{\mathrm{ECG}_u}^{(l)}
\end{equation}
where $\mathbf{W}^{(l)}$ and $\mathbf{U}^{(l)}$ are learnable parameters.

Equations \ref{eqn:tu}--\ref{eqn:te} can then be repeatedly iterated, much like is the case for any standard GNN layer. As it is possible that $G^\mathrm{ECG}$ will contain noisy edges which do not contribute to useful propagation of messages, we additionally apply DropEdge \citep{rong2020dropedge} when propagating over the ECG graph, with probability $p_{de}=0.5$.

\begin{algorithm}[h!]
\caption{Evolving Computation Graph for Graph Neural Networks: ECG-GNN}
\label{algo:method}
\SetNoFillComment
{\footnotesize 
\SetKwInput{Input}{Input}
\SetKwInput{Output}{Output}
\SetKwInput{HyperParams}{Hyper-parameters}
\SetKwInput{TrainParams}{Network Parameters}
\SetKwInput{NontrainParams}{Frozen Parameters}
\Input{
Graph $G=(V,E)$; 
Node Feature Matrix $\mathbf{X}$; Adjacency Matrix $\mathbf{A}$.
}
\HyperParams{
Value of $k$;
Drop edge probability $p_{de}$;
Number of layers $L$;
}
\Output{Predicted labels $\hat{\mathbf{y}}$}
\Begin{

\tcc{{Step 1: Extract embeddings}}
$\mathbf{H}_\mathrm{ECG}\leftarrow\gamma(\mathbf{X},\mathbf{A})$ \hfill \tcc{Embeddings stored in matrix}

\tcc{{Step 2: Construct ECG graph}}

$\mathbf{S} \leftarrow \mathbf{H}_\mathrm{ECG}\mathbf{H}_\mathrm{ECG}^\top$

\For{$u \in V$}{
    \For{$v \in V$}{
        $\hat{s}_{uv} \leftarrow s_{uv}/(\|\mathbf{h}_{\mathrm{ECG}_u}\|\|\mathbf{h}_{\mathrm{ECG}_v}\|)$ \hfill \tcc{Compute pair-wise cosine similarities}
    }
    
    $\mathcal{N}_u^\mathrm{ECG} \leftarrow \mathrm{top\text{-}}k_{v\in V} \hat{s}_{uv}$ \hfill \tcc{Compute $k$ nearest neighbours of $u$}
}
$E^\mathrm{ECG}\leftarrow \{(u, v)\ |\ u\in V\wedge v\in\mathcal{N}_u\}$ \hfill \tcc{Construct the ECG edges}

\tcc{{Step 3: Running ECG-GNN with parallel processing of $G$ and $G^\mathrm{ECG}$}}

\For{$u \in V$}{
    $\mathbf{h}^0_{u}\leftarrow \mathbf{x}_u$  \hfill \tcc{Setting initial node features}
}

\For{$l \leftarrow 1$ \KwTo $L$}{
    \tcc{Message passing propagation with the two parallel processors on $G$ and $G^\mathrm{ECG}$ respectively}
    $E^\mathrm{ECG}_{(l)}\leftarrow \mathrm{DropEdge}(E^\mathrm{ECG}, p_{de})$ \hfill \tcc{Randomly drop edges in the ECG graph}

    \For{$u \in V$}{
         $\mathbf{h}_{\mathrm{INP}_u}^{(l)} \leftarrow \phi^{(l)}_\mathrm{INP}\left( \mathbf{h}_u^{(l-1)}, \bigoplus_{(u,v) \in E} \psi_\mathrm{INP}^{(l)}\left(\mathbf{h}_u^{(l-1)}, \mathbf{h}_v^{(l-1)}\right)\right)$ \hfill \tcc{GNN on $G$}
        $\mathbf{h}_{\mathrm{ECG}_u}^{(l)}\leftarrow \phi^{(l)}_\mathrm{ECG}\left( \mathbf{h}_u^{(l-1)}, \bigoplus_{(u,v) \in E_{(l)}^\mathrm{ECG}} \psi_\mathrm{ECG}^{(l)}\left(\mathbf{h}_u^{(l-1)}, \mathbf{h}_v^{(l-1)}\right)\right)$\hfill \tcc{GNN on $G^\mathrm{ECG}$}
        
        $\mathbf{h}_{u}^{(l)}\leftarrow \mathbf{W}^{(l)} \mathbf{h}_{\mathrm{INP}_u}^{(l)} + \mathbf{U}^{(l)} \mathbf{h}_{\mathrm{ECG}_u}^{(l)}$ \hfill \tcc{Updating the node representation}
        }
    }

\tcc{Predict node labels}
\For{$u \in V$}{
    
    ${p}_{u} \leftarrow \mathrm{softmax}(\mathbf{W}^{(c)}\mathbf{h}^{(L)}_u)$
    $\hat{y}_u \leftarrow \arg \max_{c\in C}{p}_c$ \hfill \tcc{Predicted class label}
    }
\Return{$\hat{\mathbf{y}}$}
}
}
\end{algorithm}

\section{Experiments}
We evaluate the performance of ECG on five heterophilic datasets, recently-proposed by \citet{platonov2023critical}: \texttt{roman-empire}, \texttt{amazon-ratings}, \texttt{minesweeper}, \texttt{tolokers} and \texttt{questions}. All five datasets are node classification tasks, testing for varying levels of homophily in the input (\texttt{roman-empire} has the highest label informativeness), different connectivity profiles (\texttt{tolokers} is the most dense, \texttt{questions} has the lowest values of clustering coefficients)  and providing both real-world datasets (\texttt{amazon-ratings}), as well as synthetic examples (\texttt{minesweeper}).

We ran ECG as an extension on standard GNN models, choosing the ``-sep'' variant \cite{zhu2020beyond} for GAT and GT as it was noted to improve their performance consistently on these tasks \citep{platonov2023critical}. Thus, our baselines are GCN, GraphSAGE, GAT-sep and GT-sep, which we extend by modifying their computation graph as presented in Section \ref{sec:ECG}. For each ECG model, we ran three variants, depending on which weak classifier was used to select the complementary edges, $E^{\mathrm{ECG}}$: the MLP, the BGRL, or a concatenation of the output of the two.

For each of these architectures, the hyper-parameters to sweep are the number of neighbours sampled in the ECG graph, $k$ (selected from $\{3, 10, 20\}$), the edge dropout rate used on it (selected from $\{ 0., 0.5 \}$, the hidden dimension of the graph neural networks, where the one ran on the original graph $G$ always matches the one ran on $G^{\mathrm{ECG}}$ (selected from $\{ 256, 512\}$), as well as the standard choice of number of layers (selected from $\{2, 3, 4, 5\}$).  In the Appendix, we present additional information on the experiments, together with the hyper-parameters corresponding to the best validation results. 

In Table \ref{tab:results}, we show the test performance corresponding to the highest validation score among all embedding possibilites 
for each of the five datasets and for each of the four baselines. Altogether, there are 20 dataset-model combinations that ECG is tested on. We find that on 19 out of these 20 combinations (marked with arrow up in the table), using ECG improves the performance of the corresponding GNN architecture, the only exception being GraphSAGE on \texttt{amazon-ratings}.

Moreover, we observe the highest gains in performance are achieved by ECG-GCN, ranging from $1.17 \%$ to 
$10.84\%$ (absolute values) in a manner that is correlated with the homophily of the dataset. This confirms the hypothesis that, due to the aggregation function it employs, GCN is also the architecture most prone to performance changes based on the homophily of the graph. 

Additionally, we note that BGRL-based embeddings are consistently preferred for the \texttt{roman-empire} and \texttt{questions} graphs. These datasets have the lowest average degree and average local clustering -- this emphasises the need for different methods of obtaining complementary graphs, balancing connectivity and homophily aspects, as remarked in \cite{yan2021two}.

\begin{table}[t]
    \centering
    \caption{ECG performance on datasets proposed in \cite{platonov2023critical}. We report accuracy for \texttt{roman-empire} and \texttt{amazon-ratings} and ROC AUC for \texttt{minesweeper}, \texttt{tolokers}, and \texttt{questions}.}
    \label{tab:results}
    \vspace{5pt}
    \begin{tabular}{llllll}
    \toprule
    {\bf Model} &  roman-empire & amazon-ratings & minesweeper & tolokers & questions
    \\
    \midrule
MLP & $65.88 {\scriptscriptstyle\pm 0.38}$ & $45.90 {\scriptscriptstyle\pm 0.52}$ & $50.89 {\scriptscriptstyle\pm 1.39}$ & $72.95 {\scriptscriptstyle\pm 1.06}$ & $70.34 {\scriptscriptstyle\pm 0.76}$ \\

\midrule
GCN 
& $73.69 {\scriptscriptstyle\pm 0.74}$ & $48.70 {\scriptscriptstyle\pm 0.63}$ & $89.75 {\scriptscriptstyle\pm 0.52}$ & $83.64 {\scriptscriptstyle\pm 0.67}$ & $76.09 {\scriptscriptstyle\pm 1.27}$ \\

ECG-GCN & $84.53 {\scriptscriptstyle\pm 0.26}$ ($\uparrow$) & {$51.12 {\scriptscriptstyle\pm 0.38}$} ($\uparrow$) & {$92.63 {\scriptscriptstyle\pm 0.10}$} ($\uparrow$) & $\textbf{84.81} {\scriptscriptstyle\pm 0.25}$ ($\uparrow$) & $77.50 {\scriptscriptstyle\pm 0.35}$ ($\uparrow$) \\

\midrule

SAGE 
& $85.74 {\scriptscriptstyle\pm 0.67}$ & $53.63 {\scriptscriptstyle\pm 0.39}$ & $93.51 {\scriptscriptstyle\pm 0.57}$ & $82.43 {\scriptscriptstyle\pm 0.44}$ & $76.44 {\scriptscriptstyle\pm 0.62}$ \\

ECG-SAGE & {$87.88 {\scriptscriptstyle\pm 0.25}$}  ($\uparrow$)	& $53.45 {\scriptscriptstyle\pm 0.27}$  ($\downarrow$)	& {$94.11 {\scriptscriptstyle\pm 0.07}$}  ($\uparrow$)	& {$82.61 {\scriptscriptstyle\pm 0.29}$}  ($\uparrow$)	& {$77.23 {\scriptscriptstyle\pm 0.36}$}  ($\uparrow$) \\

\midrule

GAT-sep 
& {$88.75 {\scriptscriptstyle\pm 0.41}$}  & {$52.70 {\scriptscriptstyle\pm 0.62}$} & {$93.91 {\scriptscriptstyle\pm 0.35}$} & {$83.78 {\scriptscriptstyle\pm 0.43}$} & $76.79 {\scriptscriptstyle\pm 0.71}$ \\

ECG-GAT-sep & $\textbf{89.62} {\scriptscriptstyle\pm 0.18}$  ($\uparrow$) & $\textbf{53.65} {\scriptscriptstyle\pm 0.39}$  ($\uparrow$) & $\textbf{94.52} {\scriptscriptstyle\pm 0.20}$  ($\uparrow$) & {$84.23 {\scriptscriptstyle\pm 0.25}$}  ($\uparrow$) & {$77.38 {\scriptscriptstyle\pm 0.18}$}  ($\uparrow$) \\

\midrule 
GT-sep 
& {$87.32 {\scriptscriptstyle\pm 0.39}$} & $52.18 {\scriptscriptstyle\pm 0.80}$ & {$92.29 {\scriptscriptstyle\pm 0.47}$} & $82.52 {\scriptscriptstyle\pm 0.92}$ & {$78.05 {\scriptscriptstyle\pm 0.93}$} 
\\
ECG-GT-sep & $89.56 {\scriptscriptstyle\pm 0.16}$  ($\uparrow$) & $53.25 {\scriptscriptstyle\pm 0.39}$ ($\uparrow$)  & $93.62 {\scriptscriptstyle\pm 0.27}$  ($\uparrow$) & $84.00 {\scriptscriptstyle\pm 0.24}$ ($\uparrow$) & $78.12 {\scriptscriptstyle\pm 0.32}$ ($\uparrow$) \\

\midrule
H$_2$GCN & $60.11 {\scriptscriptstyle\pm 0.52}$ & $36.47 {\scriptscriptstyle\pm 0.23}$  &$89.71 {\scriptscriptstyle\pm 0.31}$ & $73.35 {\scriptscriptstyle\pm 1.01}$ & $63.59 {\scriptscriptstyle\pm 1.46}$ \\
CPGNN &$63.96 {\scriptscriptstyle\pm 0.62}$ & $39.79 {\scriptscriptstyle\pm 0.77}$ & $52.03 {\scriptscriptstyle\pm 5.46}$ & $73.36 {\scriptscriptstyle\pm 1.01}$ & $65.96 {\scriptscriptstyle\pm 1.95}$  \\
GPR-GNN & $64.85 {\scriptscriptstyle\pm 0.27}$ & $44.88 {\scriptscriptstyle\pm 0.34}$ & $86.24 {\scriptscriptstyle\pm 0.61}$ & $72.94 {\scriptscriptstyle\pm 0.97}$ & $55.48 {\scriptscriptstyle\pm 0.91}$ \\
FSGNN & $79.92 {\scriptscriptstyle\pm 0.56}$ & $52.74 {\scriptscriptstyle\pm 0.83}$ & $90.08 {\scriptscriptstyle\pm 0.70}$ & $82.76 {\scriptscriptstyle\pm 0.61}$ & $\textbf{78.86} {\scriptscriptstyle\pm 0.92}$ \\
GloGNN & $59.63 {\scriptscriptstyle\pm 0.69}$ & $36.89 {\scriptscriptstyle\pm 0.14}$ & $51.08 {\scriptscriptstyle\pm 1.23}$ & $73.39 {\scriptscriptstyle\pm 1.17}$ & $65.74 {\scriptscriptstyle\pm 1.19}$ \\
FAGCN & $65.22 {\scriptscriptstyle\pm 0.56}$ & $44.12 {\scriptscriptstyle\pm 0.30}$ & $88.17 {\scriptscriptstyle\pm 0.73}$ & $77.75 {\scriptscriptstyle\pm 1.05}$ & $77.24 {\scriptscriptstyle\pm 1.26}$ \\
GBK-GNN & $74.57 {\scriptscriptstyle\pm 0.47}$ & $45.98 {\scriptscriptstyle\pm 0.71}$ & $90.85 {\scriptscriptstyle\pm 0.58}$ & $81.01 {\scriptscriptstyle\pm 0.67}$ & $74.47 {\scriptscriptstyle\pm 0.86}$ \\
JacobiConv & $71.14 {\scriptscriptstyle\pm 0.42}$ & $43.55 {\scriptscriptstyle\pm 0.48}$ & $89.66 {\scriptscriptstyle\pm 0.40}$ & $68.66 {\scriptscriptstyle\pm 0.65}$ & $73.88 {\scriptscriptstyle\pm 1.16}$ \\
    \bottomrule
    \end{tabular}
\end{table}

\begin{figure}
    \centering
    \includegraphics[width=.48\linewidth]{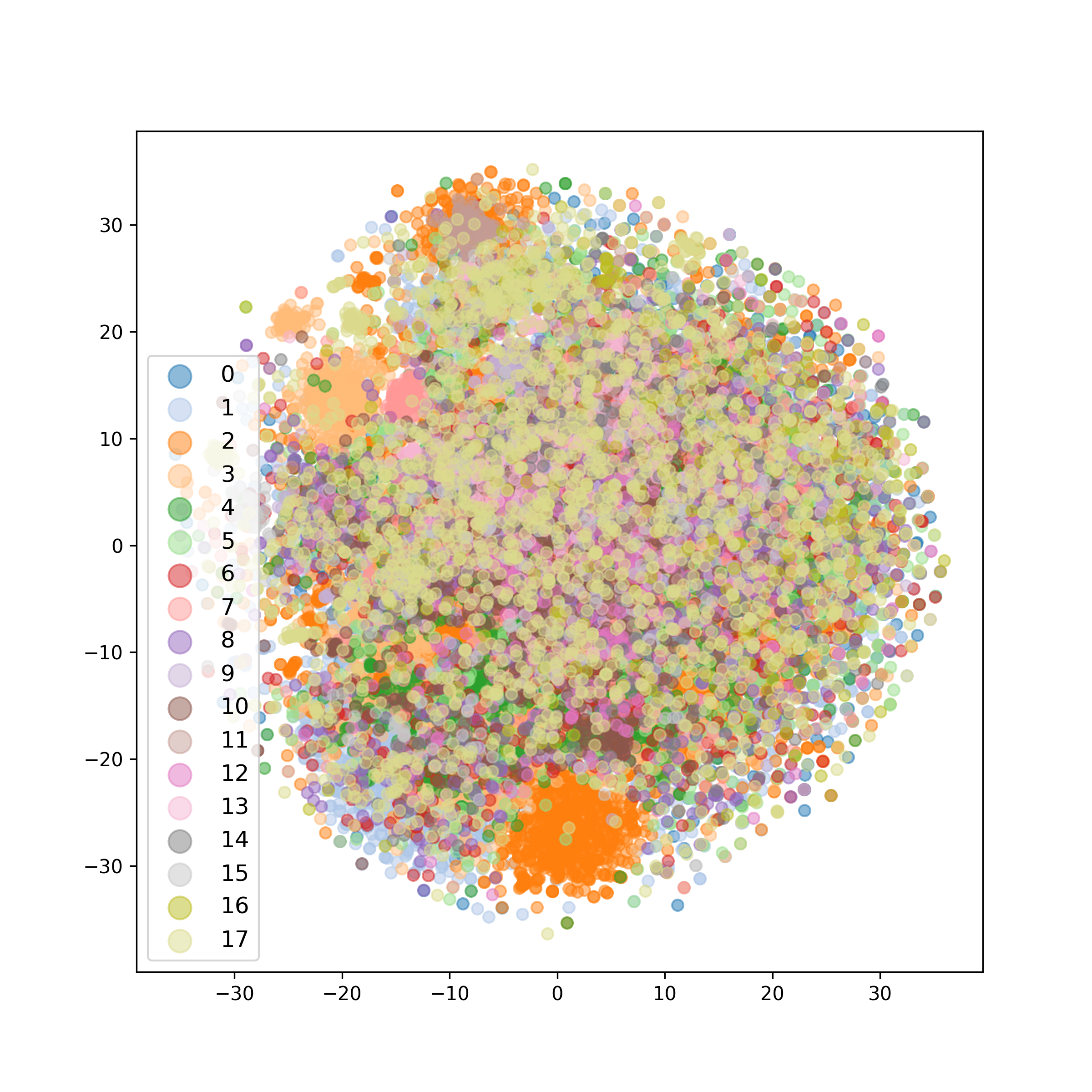}
    \includegraphics[width=.48\linewidth]{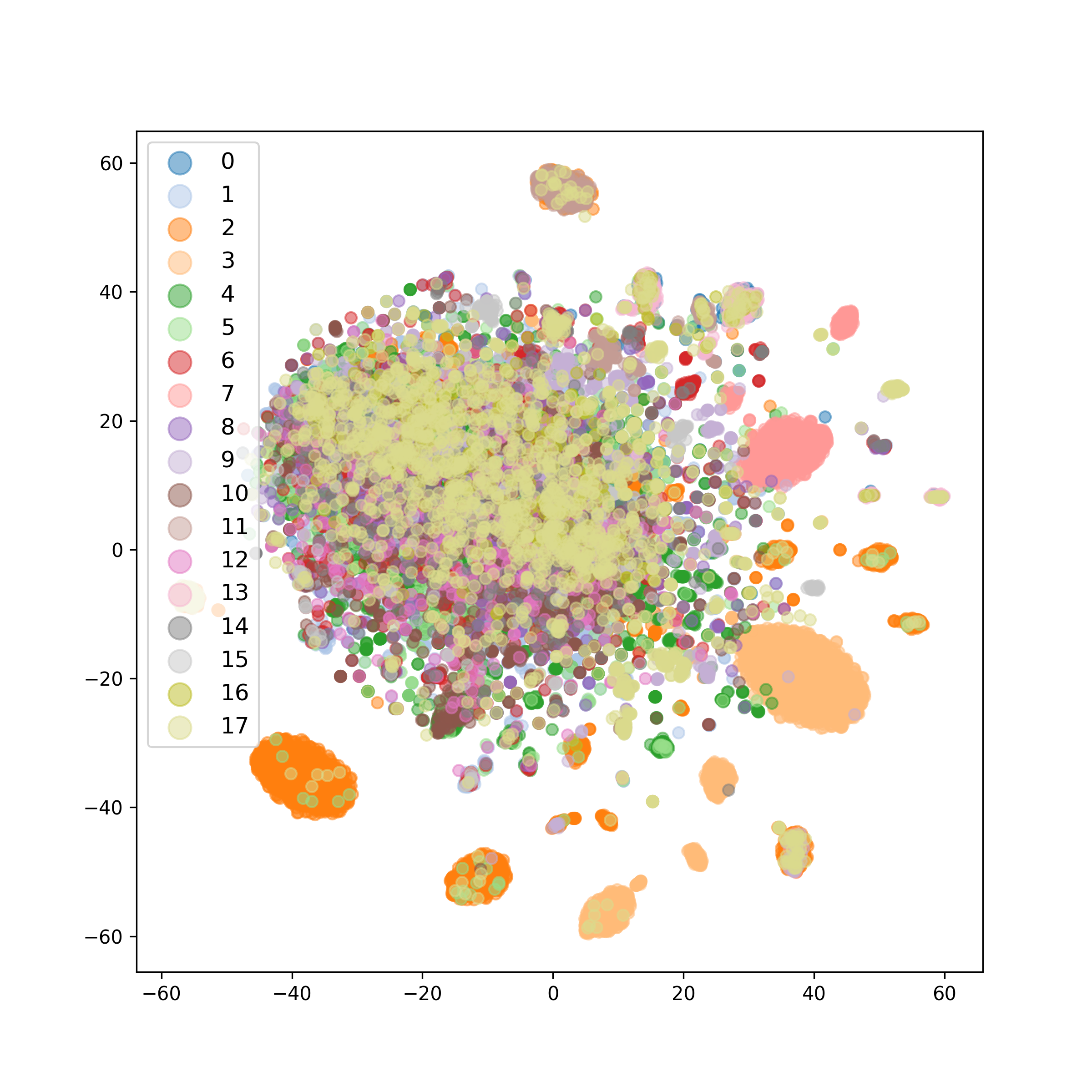}
    \caption{For \texttt{roman-empire}, we use a random GCN layer to obtain node embeddings based on the original graph $G$ (left) or from the complementary graph $G^{\mathrm{ECG}}$ (right). The colours correspond to the ground-truth labels of the nodes.}
    \label{fig:tsne_g_vs_MLPgc}
\end{figure}

\subsection{Qualitative studies}

In Table \ref{tab:statistics}, we analyse the properties of the complementary graphs $G^{\mathrm{ECG}}$ with $k=3$ nearest neighbours. We note that this represents the graph used by ECG-GCN, which preferred lower values of $k$, while the optimal values of $k$ for SAGE, GAT-sep and GT-sep were on the higher end, varying between $3$, $10$ and $20$ depending on the dataset.

We observe that MLP-ECG confirms our hypothesis: taking the edges corresponding to the pairs of nodes marked as most similar by the ResNet results in a graph $G^{\mathrm{ECG}}$ with high homophily, especially compared to the original input graph. 
It is important to note that all of our MLP-ECG graphs were obtained with a relatively shallow ResNet, which, as it can be seen in Table \ref{tab:results} lacks in performance compared to the graph-based methods. However, our method's success in conjunction with GNNs shows that even a weak classifier can be used to generate homophilic graphs that can improve performance when used to complement the information provided by the given input data.

In Figure \ref{fig:tsne_g_vs_MLPgc}, we also verify how predictive of the node classes the graph topology is when obtained from the original data compared to when we build a complementary graph $G^{\mathrm{ECG}}$. More precisely, we first build the graph $G^{\mathrm{ECG}}$ as presented in Step 1 of Algorithm \ref{algo:method}, using pre-trained MLP embeddings. Then we use a randomly initialised GCN to compute node embeddings on the input graph $G$, as well as on $G^{\mathrm{ECG}}$. We visualise these two sets of node embeddings using  t-SNE \cite{van2008visualizing} by projecting to a 2D space, attributing the colour of each point based on the node's ground truth label. We can observe that using the $G^{\mathrm{ECG}}$ topology leads to more distinguishable clusters corresponding to the classes even without any training, thus supporting the enhancements in performance when building ECG-GNN.

\begin{table}[t]
    \centering
        \caption{Statistics of the original heterophilous graphs and of the evolutionary computation graph obtained from MLP and BGRL.}
    \label{tab:statistics}
\vspace{5pt}
    \begin{footnotesize}
    \begin{tabular}{lcccccc}
    \toprule
    & roman-empire & amazon-ratings & minesweeper & tolokers & questions
    \\
    \midrule
    edges & $32,927$ & $93,050$ & $39,402$ & $519,000$ & $153,540$
    \\
    edge homophily & $0.05$ & $0.38$ & $0.68$ & $0.59$ & $0.84$
    \\
    adjusted homophily & $-0.05$ & $0.14$ & $0.01$ & $0.09$ & $0.02$
    \\
    LI & $0.11$ & $0.04$ & $0.00$ & $0.01$ & $0.00$
    \\
    \midrule
    ECG($k=3$) edges & $67,986$ & $73,476$ & $30,000$ & $35,274$ & $146,763$
    \\
    MLP-ECG edge homophily & $0.73$ & $0.66$ & $0.79$ & $0.79$ & $0.97$
    \\
    MLP-ECG adjusted homophily & $0.7$ & $0.53$ & $0.33$ & $0.4$ & $0.41$
    \\
    MLP-ECG LI & $0.65$ & $0.33$ & $0.16$ & $0.19$ & $0.28$
    \\
    \midrule

    BGRL-ECG edge homophily & $0.16$ & $0.3$ & $0.68$ & $0.6$ & $0.93$
    \\
    BGRL-ECG adjusted homophily & $0.06$ & $0.02$ & $0.12$ & $0.08$ & $0.01$
    \\
    BGRL-ECG LI & $0.1$ & $0.03$ & $0.05$ & $0.03$ & $0.03$ \\
    \bottomrule
    \end{tabular}
    \end{footnotesize}
\end{table}

\section{Related work}

Many specialised architectures have been proposed to tackle GNNs limitations in modeling heterophilic data. 
H$_2$GCN \cite{zhu2020beyond} proposes separation of ego and neighbour embeddings, using higher-order neighbourhoods and combining representations from intermediate layers. 
CPGNN \cite{zhu2020graph} learns a compatibility matrix to explicitly integrate information about label correlations and uses it to modulate messages. 
Similarly, GGCN \cite{yan2021two} modifies GCN through degree corrections and signed messages, based on an insight linking heterophily and oversmoothing.
FAGCN \cite{bo2021beyond} uses a self-gating mechanism to adaptively integrate low-frequency signals, high-frequency signals and raw features and ACM-GNN \cite{luan2022revisiting} extends it to enable adaptive channel mixing node-wise.
GPRGNN \cite{chien2021adaptive} learns Generalized PageRank weights that adjust to node label patterns.  
FSGNN \cite{maurya2022simplifying} proposes Feature Selection GNN which separates node feature aggregation from the depth of the GNN through multiplication the node features with different powers and transformations of the adjacency matrix and uses a softmax to select the relevant features. GloGNN \cite{li2022finding} leverages global nodes to aggreagte information, while GBK-GNN \cite{du2022gbk} uses bi-kernel feature transformation.

Most relevant to ECG-GNN could be considered GeomGCN \cite{pei2020geom} and the work of \citet{suresh2021breaking}. The former uses network embedding methods to find neighbouring nodes in the corresponding latent space, to be then aggregated with the neighbours in the original graph, over which a two-level aggregation is then performed. Similarly, \cite{suresh2021breaking} modifies the computation graph by computing similarity of degree sequences for different numbers of hops. However, in both cases, the input node features and labels are not used, making it prone to inaccuracies in the graph structure.

It was recently pointed \cite{platonov2023critical} that the standard datasets on which these models were tested, such as Squirrel, Chameleon, Cornell, Texas and Wisconsin, had considerable drawbacks: high number of duplicated nodes, highly imbalanced classes and lack of diversity in setups considered. In fact, when evaluated on their newly proposed heterophilic suite, it was noted that most specialised architectures are outperformed by their standard counter-parts such as GCN, SAGE, GAT and GT, with only the separation of ego and neighbour embeddings from \cite{zhu2020beyond} maintaining an advantage. 

\section{Limitations and further work}
We observe that our complementary graph is constructed based on two sources of homophily: a pretrained ResNet model that relies on provided training labels, and a pretrained self-supervised graph module, BGRL, which depends solely on the graph structure and node features without any labels. In cases where neither of these two approaches generates a graph with a satisfactory level of homophily and advantageous connectivity, the complementary graph may struggle to enhance the overall model performance, as evidenced by the relatively smaller gains observed in the \texttt{amazon-ratings} dataset.

In such cases, it may be beneficial to explore additional sources for obtaining embeddings. Furthermore, there is potential for improvements in leveraging the complementary information, such as separately using three distinct graphs: the original graph, the MLP-ECG graph, and the BGRL-ECG graph. This approach might prove to be more effective in incorporating the different types of information, rather than relying solely on the projection of concatenated embeddings.

Finally, it is worth noting that while this method enhances the performance of standard graph neural networks, it can also be applied to specialised architectures specifically designed to improve performance on heterophilic data. These two approaches are independent and can in principle be combined to further boost a model's capabilities. By integrating the complementary graph construction into specialized architectures, we could leverage the benefits of both techniques and potentially achieve even better results when dealing with heterogeneous data. As our paper focuses on the effects of modulating the graph structure using weak classifiers, we apply only commonly-used GNN layers, and leave explorations of this kind to future work.

\section{Conclusions}

We present Evolving Computation Graphs for graph neural networks, ECG-GNN. This is a two-phase method focused on improving the performance of standard GNNs on heterophilous data. Firstly, it builds an evolved computation graph, formed from the original input and a complementary set of edges determined by a weak classifier. Then, the two components of this computation graph are modelled in parallel by two GNNs processors, and projected to the same embeddings space after each propagation layer. This simple and elegant extension of existing graph neural networks proves to be very effective -- for four models considered on five diverse heterophilic datasets, the ECG-GNN enhances the performance in $95\%$ of the cases.

\bibliographystyle{plainnat}
\bibliography{references}

\newpage

\appendix

\section{Training information}\label{app:training}

We used the same experimental setup as presented in \citet{platonov2023critical}. Results are aggregated over ten random splits of the data, with each run taking $50\%$ of the nodes for training, $25\%$ for validation, and $25\%$ for testing. The following hyperparameters are tuned for all models and baselines, using the average validation performance across the splits:
\begin{itemize}
    \item Number of GNN/ResNet layers, $L\in\{1, 2, 3, 4, 5\}$.
    \item The dimensionality of the GNN/ResNet's latent embeddings, $d\in\{256,512,1024\}$.
\end{itemize}
Additionally, for the ECG models only, the following hyperparameters were swept:
\begin{itemize}
    \item Embeddings used by ECG, $\tau\in\{\text{MLP},\text{BGRL},\text{MLPBGRL},\text{MLP}\rightarrow\text{GNN}\}$, referring to:
    \begin{itemize}
        \item[\em MLP:] Using the embeddings from a pre-trained ResNet;
        \item[\em BGRL:] Using the embeddings from a pre-trained BGRL model;
        \item[\em MLPBGRL:] Using the normalised concatenation of the ResNet and BGRL embeddings;
        \item[\em MLP$\rightarrow$GNN:] Using the embeddings from a pre-trained MLP-ECG model of the same type.
    \end{itemize}
    Note that, for methods requiring access to labels (such as MLP), a separate set of embeddings is computed for every dataset split (to avoid test data contamination). For self-supervised methods like BGRL, no labels are used, and hence a single set of embeddings is produced for all experiments.
    \item The number of neighbours sampled per node, $k\in\{3,10,20\}$.
    \item The DropEdge rate, $p_{de}\in\{0.0, 0.5\}$.
\end{itemize}

The model configuration with the best-performing average validation performance is then evaluated on the corresponding test splits, producing the aggregated performances reported in Table \ref{tab:results}.

The best-performing hyperparameters for each model type on each dataset are given in Table \ref{tab:best_hparams}. Each individual experiment has been executed on a single NVIDIA Tesla P100 GPU, and the longest training time allocated to an individual experiment has been six hours (on the \texttt{questions} dataset).

For convenience, and to assess the relative benefits of various ECG embedding sources, we provide in Table \ref{tab:additional_results} an expanded version of \ref{tab:results}, showing the test performance obtained by the tuned version of each ECG variant, for every embedding type.

For additional information, the anonymised code can be found at \url{https://anonymous.4open.science/r/evolving_computation_graphs-97B7/}.

\begin{table}[t]
\centering
    \caption{The best-performing hyperparameters for each GNN propagation rule in our experiments. The only experiment where the baseline model outperforms ECG is the SAGE propagation layer on \texttt{amazon-ratings}; hence, the hyperparameters $k$ and $p_{de}$ are irrelevant.}
    \label{tab:best_hparams}
    \vspace{5pt}
    \begin{tabular}{lccccc}
    \toprule
    & roman-empire & amazon-ratings & minesweeper & tolokers & questions
    \\
    \midrule
{\bf ResNet}\\
$L$ & $2$ & $1$ & $5$ & $5$ & $1$\\
$d$ & $512$ & $512$ & $512$ & $512$ & $512$\\
\midrule
{\bf GCN} \\
$L$ & $5$ & $2$ & $4$ & $4$ & $3$\\
$d$ & $512$ & $512$ & $256$ & $512$ & $256$\\
$\tau$ & MLPBGRL & MLP$\rightarrow$GNN & MLP & MLP & BGRL\\
$k$ & $3$ & $3$ & $3$ & $3$ & $3$\\
$p_{de}$ & $0.5$ & $0.5$ & $0.5$ & $0.5$ & $0.0$\\
\midrule
{\bf SAGE} \\
$L$ & $5$ & $2$ & $5$ & $4$ & $5$\\
$d$ & $512$ & $1024$ & $256$ & $256$ & $256$\\
$\tau$ & BGRL & Baseline & BGRL & BGRL & BGRL\\
$k$ & $10$ & --- & $20$ & $20$ & $10$\\
$p_{de}$ & $0.5$ & --- & $0.5$ & $0.5$ & $0.0$\\
\midrule
{\bf GAT-sep} \\
$L$ & $5$ & $2$ & $5$ & $5$ & $4$\\
$d$ & $512$ & $512$ & $256$ & $256$ & $256$\\
$\tau$ & BGRL & MLP$\rightarrow$GNN & MLP & BGRL & BGRL\\
$k$ & $10$ & $3$ & $20$ & $20$ & $10$\\
$p_{de}$ & $0.5$ & $0.5$ & $0.5$ & $0.5$ & $0.5$\\
\midrule
{\bf GT-sep} \\
$L$ & $5$ & $2$ & $5$ & $5$ & $4$\\
$d$ & $512$ & $512$ & $256$ & $256$ & $256$\\
$\tau$ & BGRL & MLP$\rightarrow$GNN & MLP & MLPBGRL & BGRL\\
$k$ & $20$ & $3$ & $20$ & $20$ & $10$\\
$p_{de}$ & $0.5$ & $0.5$ & $0.0$ & $0.0$ & $0.5$\\
\bottomrule
\end{tabular}
\end{table}

\begin{table}[t]
    \centering
    \caption{Detailed breakdown of model performance on the datasets proposed by \citet{platonov2023critical}. ResNet, GCN, SAGE, GAT-sep and GT-sep are the baselines, while all the other models are variants of ECG. Red marks the best performance on each dataset for each of the considered GNN architectures and the corresponding ECGs.
    Accuracy is reported for \texttt{roman-empire} and \texttt{amazon-ratings}, and ROC AUC is reported for \texttt{minesweeper}, \texttt{tolokers}, and \texttt{questions}.}
    \label{tab:additional_results}
    \vspace{5pt}
    \begin{tabular}{lccccc}
    \toprule
    & roman-empire & amazon-ratings & minesweeper & tolokers & questions
    \\
    \midrule
ResNet & $65.88 \pm 0.38$ & $45.90 \pm 0.52$ & $50.89 \pm 1.39$ & $72.95 \pm 1.06$ & $70.34 \pm 0.76$ \\
\midrule
GCN & $73.69 \pm 0.74$ & $48.70 \pm 0.63$ & $89.75 \pm 0.52$ & $83.64 \pm 0.67$ & $76.09 \pm 1.27$ \\

MLP-ECG-GCN & $83.55 \pm 0.39$ & $50.99 \pm 0.64$ & \textcolor{red}{$92.63 \pm 0.10$} & \textcolor{red}{$84.81 \pm 0.25$} & $76.25 \pm 0.59$ \\
BGRL-ECG-GCN & $80.59 \pm 0.48$	& $48.99 \pm 0.28$ 	& $92.35 \pm 0.10$	& $84.25 \pm 0.22$	& \textcolor{red}{$77.50 \pm 0.35$} \\
MLPBGRL-ECG-GCN & \textcolor{red}{$84.53 \pm 0.26$}	& $50.11 \pm 0.60$	& $92.47 \pm 0.50$	& $84.73 \pm 0.23$	& $77.32 \pm 0.31$ \\
MLP->GNN-ECG-GCN & $84.39 \pm 0.22$	& \textcolor{red}{$51.12 \pm 0.38$}	& $92.56 \pm 0.23$	& $84.35 \pm 0.31$ & $75.16 \pm 0.87$ \\

\midrule

SAGE & $85.74 \pm 0.67$ & \textcolor{red}{$53.63 \pm 0.39$} & {$93.51 \pm 0.57$} & $82.43 \pm 0.44$ & $76.44 \pm 0.62$ \\

MLP-ECG-SAGE & $85.82 \pm 0.62$	& $53.32 \pm 0.39$	& $94.10 \pm 0.08$	& $82.60 \pm 0.23$	& $76.13 \pm 0.41$ \\

BGRL-ECG-SAGE & \textcolor{red}{$87.88 \pm 0.25$}	& $53.12 \pm 0.32$	& \textcolor{red}{$94.11 \pm 0.07$} &	\textcolor{red}{$82.61 \pm 0.29$} & \textcolor{red}{$77.23 \pm 0.36$} \\

MLPBGRL-ECG-SAGE & $86.50 \pm 0.34$	& $52.34 \pm 0.92$	& $94.01 \pm 0.07$	& $82.55 \pm 0.18$	& $76.55 \pm 0.33$ \\
MLP->GNN-ECG-SAGE & $85.94 \pm 0.57$	& $53.45 \pm 0.27$	& $93.77 \pm 0.12$	& $82.52 \pm 0.22$ & $75.53 \pm 0.64$\\	

\midrule

GAT-sep & {$88.75 \pm 0.41$} & {$52.70 \pm 0.62$} & {$93.91 \pm 0.35$} & {$83.78 \pm 0.43$} & $76.79 \pm 0.71$ \\

MLP-ECG-GAT-sep & $88.22 \pm 0.36$	& $52.98 \pm 0.30$	& \textcolor{red}{$94.52 \pm 0.20$}	& $83.91 \pm 0.32$	& $77.30 \pm 0.47$ \\
BGRL-ECG-GAT-sep & \textcolor{red}{$89.62 \pm 0.18$}	& $52.20 \pm 0.57$	& $94.24 \pm 0.15$	& \textcolor{red}{$84.23 \pm 0.25$}	& \textcolor{red}{$77.38 \pm 0.18$} \\
MLPBGRL-GAT-sep & $88.73 \pm 0.37$	& $51.06 \pm 0.73$	& $94.39 \pm 0.20$	& $84.11 \pm 0.23$	& $76.97 \pm 0.45$ \\
MLP->GNN-ECG-GAT-sep & $88.04 \pm 0.32$	& \textcolor{red}{$53.65 \pm 0.39$}	& $93.97 \pm 0.19$ 	& $83.75 \pm 0.30$ & $75.61 \pm 0.74$\\

\midrule
GT-sep & {$87.32 \pm 0.39$} & $52.18 \pm 0.80$ & {$92.29 \pm 0.47$} & $82.52 \pm 0.92$ & {$78.05 \pm 0.93$} \\
MLP-ECG-GT-sep & $88.56 \pm 0.35$	& {$52.68 \pm 0.65$}	& \textcolor{red}{$93.62 \pm 0.27$}	& {$83.65 \pm 0.29$} & $ 77.82 \pm 0.43 $\\	
BGRL-ECG-GT-sep & \textcolor{red}{$89.56 \pm 0.16$}	& $52.37 \pm 0.30$	& $93.55 \pm 0.18$	& $82.97 \pm 0.26$	& \textcolor{red}{$78.12 \pm 0.32$} \\
MLPBGRL-GT-sep &  $88.70 \pm 0.30$	& $52.29 \pm 0.60$	& $93.52 \pm 0.25$	& \textcolor{red}{$84.00 \pm 0.24$}	& $77.85 \pm 0.45$\\
MLP->GNN-ECG-GT-sep & $88.62 \pm 0.46$	& \textcolor{red}{$53.25 \pm 0.39$}	& $92.69 \pm 0.34$	& $83.41 \pm 0.44$	& $75.50 \pm 1.13$ \\
    \bottomrule
    \end{tabular}
\end{table}

\end{document}